\documentclass[preprint,review,12pt,authoryear,3p]{elsarticle}

\usepackage{microtype}
\usepackage{amsmath,amssymb}
\usepackage{graphicx}
\usepackage{booktabs}
\usepackage{array}
\usepackage{multirow}
\usepackage{caption}
\usepackage{subcaption}
\usepackage{float}
\usepackage{enumitem}
\usepackage[hidelinks]{hyperref}
\usepackage{xcolor}
\usepackage{adjustbox}

\journal{Neural Networks}
\biboptions{authoryear,round}
\graphicspath{{figures/}}

\newcommand{\confacc}{\mathrm{CA}_{\mathrm{conflict}}}
\newcommand{\texchoice}{\mathrm{TC}_{\mathrm{conflict}}}
\newcommand{\ece}{\mathrm{ECE}}
\newcommand{\sscore}{\mathrm{SS}}
\newcommand{\ft}{f_{\mathrm{target}}}

\begin{document}
\begin{frontmatter}

\title{Early Cue Precision Shapes Visual Shortcut Learning in Controlled Cue-Manipulation Benchmarks}

\author[aff1]{Chanho Park}
\author[aff1]{Woochan Lee}
\author[aff1]{Jangyeong Oh}
\author[aff1]{Geongho Gong}
\author[aff1]{Minsu Kim}
\author[aff1]{Yeachan Kwak}
\author[aff1]{Seongim Choi\corref{cor1}}
\ead{schoi1@gist.ac.kr}
\address[aff1]{Gwangju Institute of Science and Technology, Gwangju, Republic of Korea}
\cortext[cor1]{Corresponding author.}

\begin{abstract}
Visual classifiers can achieve high matched-distribution accuracy while relying on low-level cues that fail under conflict or suppression. We test whether this failure is shaped by \emph{early cue precision}: the reliability with which a low-level cue predicts the label during early learning or downstream probe fitting. Across synthetic shape-texture tasks, sequential digit training, a 10-class frozen-representation audit, and a CIFAR-10 natural-image-based texture-overlay benchmark, we manipulate object-texture match probability and evaluate matched-ID accuracy, conflict accuracy, texture-choice rate, and suppression behavior. Degraded-but-predictive input does not substitute for cue decorrelation. In 10-class digit probes, conflict accuracy drops from 0.589 under chance-like cue precision to 0.005 under target-perfect texture. In CIFAR-10 frozen probes, conflict accuracy drops from 0.569 to 0.114, while texture choice rises from 0.049 to 0.855; this ordering persists across texture-overlay strengths $\alpha\in\{0.15,0.25,0.35,0.50\}$. End-to-end CIFAR-10 training shows that low early cue precision improves pre-target conflict behavior, but shortcut-rich fine-tuning can rapidly overwrite this benefit. Cue decorrelation must therefore be maintained during downstream adaptation rather than treated as a one-time inoculation.
\end{abstract}


\end{frontmatter}

\section{Introduction}

This study began from a biological intuition. Predictive-processing accounts of perception emphasize how systems weight sensory evidence and prior expectations under uncertainty \citep{sterzer2018predictive}, and work on congenital or early blindness has motivated hypotheses about altered developmental reliance on visual evidence and cortical reorganization \citep{silverstein2013blindness,leivada2014schizophrenia,morgan2018congenital}. We do not model blindness, psychosis, schizophrenia, or human neurodevelopment. Instead, this motivation leads to a machine-learning question: does robustness depend on the absence or degradation of visual input, or on preventing low-level visual cues from becoming excessively reliable early in learning?

Modern visual recognition systems can perform well on matched test data while relying on evidence that does not correspond to the causal structure of the task. Texture, color, local artifacts, and background context can be reliable in the training distribution but fail when object identity and low-level cues are decoupled. This phenomenon is a central instance of shortcut learning \citep{geirhos2020shortcut}. Texture-shape conflict studies have shown that ImageNet-trained convolutional networks can prefer texture over global shape, and that increasing shape bias can improve robustness \citep{geirhos2019imagenet}. Related work on dataset bias, non-robust features, spurious correlations, common corruptions, natural adversarial examples, and bias-controlled object recognition further shows that high benchmark accuracy does not guarantee stable recognition under altered evidence \citep{torralba2011unbiased,ilyas2019adversarial,sagawa2020distributionally,hendrycks2019benchmarking,hendrycks2021natural,barbu2019objectnet}.

One interpretation is architectural: convolutional networks are texture-biased, whereas transformers, self-supervised models, or vision-language encoders may be more shape-oriented. Architecture and pretraining do matter \citep{dosovitskiy2021vit,liu2022convnet,liu2021swin,radford2021clip,oquab2024dinov2}. However, shortcut acquisition is also a learning-history problem. If a low-level cue reliably predicts the label early in training or during downstream probe fitting, many model classes may learn a decision rule that treats that cue as highly informative. Conversely, if the same cue is visible but unreliable, or if structural information is privileged before the cue becomes predictive, later decisions may rely less on the shortcut.

We call this the \emph{early cue-precision hypothesis}. In our experiments, cue precision is manipulated by the match probability $p=P(t=y)$ between the low-level texture label $t$ and the target label $y$. Equivalently, under the symmetric $K$-class construction used here, cue precision can be expressed as the cue-label mutual information
\begin{equation}
I(Y;T)=\log_2K+p\log_2p+(1-p)\log_2\frac{1-p}{K-1},
\label{eq:mi}
\end{equation}
with $I(Y;T)=0$ at chance-level matching ($p=1/K$) and $I(Y;T)=\log_2K$ when texture perfectly predicts the label. Thus, ``precision'' in this paper means operational cue-label reliability, not a claim about neural precision coding. The hypothesis is not that degraded visual input improves robustness. It is that shortcut reliance is reduced when low-level cues fail to acquire excessive early predictive reliability. A degraded input stream can still contain a highly predictive shortcut, so degradation and cue decorrelation are computationally distinct interventions.

\paragraph{Contributions} This paper makes five contributions. First, it formulates visual shortcut learning as an early cue-reliability problem rather than a CNN-only architectural defect. Second, it separates degraded-but-predictive input from cue-decorrelation curricula. Third, it introduces a unified protocol with matched-ID, conflict, texture-suppressed, and object-suppressed evaluation. Fourth, it tests the hypothesis in four layers: synthetic shape-texture curricula, true sequential digit learning, a 10-class frozen-representation audit, and a CIFAR-10 natural-image-based texture-overlay benchmark with frozen and end-to-end variants. Fifth, it adds color, PCA, probe-regularization, paired-permutation, bootstrap, leave-one-model, accuracy-matched, and suppression controls. We separate \emph{learning-path evidence}, where training history is manipulated, from \emph{representation-audit evidence}, where cue precision is manipulated only during downstream probe fitting on frozen encoders.

\begin{figure}[H]
\centering
\includegraphics[width=0.98\textwidth]{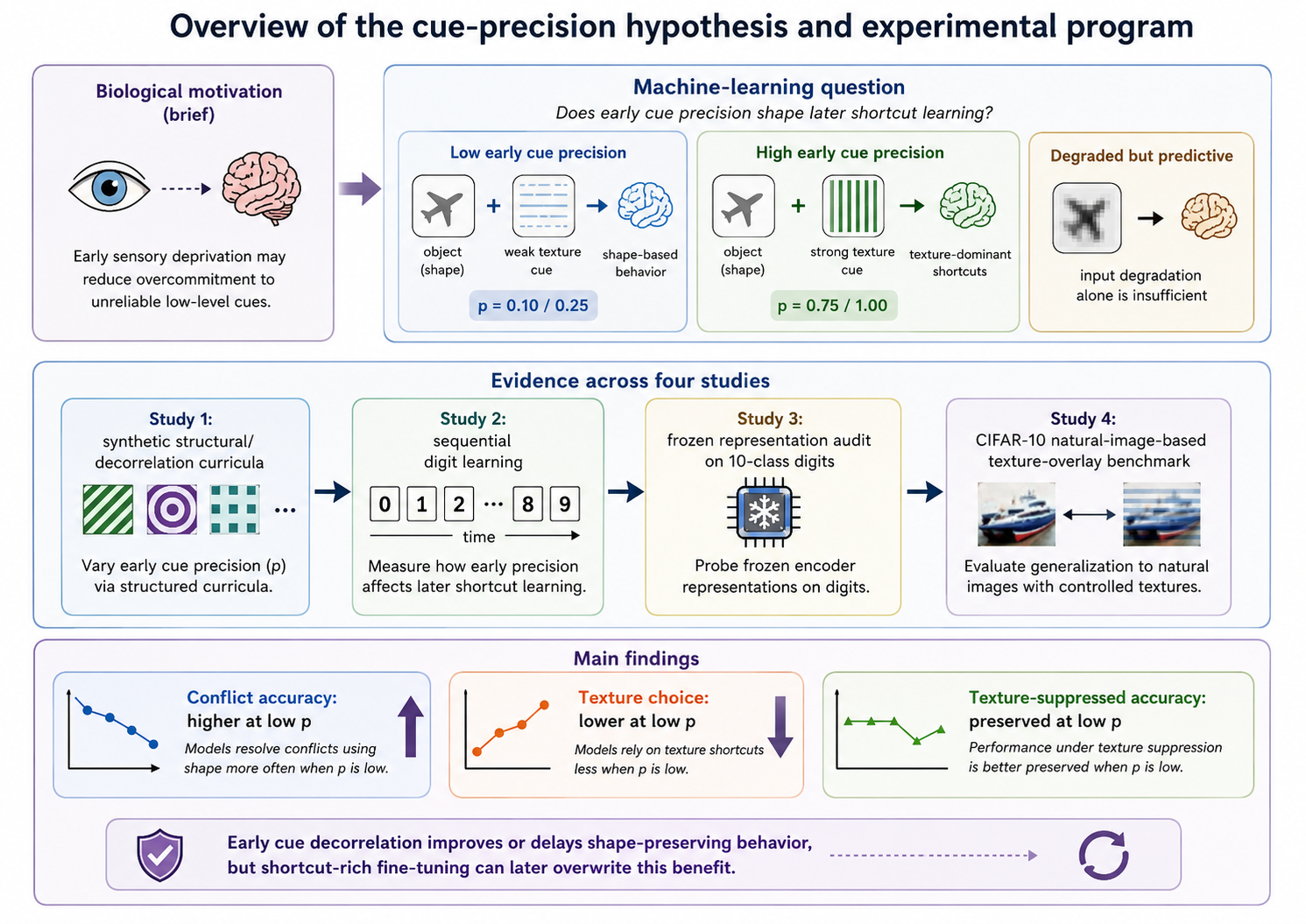}
\caption{Overview of the cue-precision hypothesis and experimental program. The biological panel is included only as a brief motivation; the technical question is whether early cue-label reliability shapes later shortcut reliance. The middle row summarizes the four experimental layers, and the bottom row summarizes the primary outcomes used throughout the paper.}
\label{fig:overview-program}
\end{figure}

\section{Related Work}

\subsection{Shortcut learning and texture bias}

The texture-shape cue-conflict paradigm separated object shape responses from texture responses in image classifiers \citep{geirhos2019imagenet}. Subsequent work framed such failures as shortcut learning: a model uses a rule that is predictive under the training distribution but unstable under shift \citep{geirhos2020shortcut}. Analyses of texture-bias origins suggest that data and augmentation can be as important as architecture \citep{hermann2020origins}. This supports our focus on the reliability and timing of training cues.

Shortcut learning also connects to spurious correlations and group robustness. Distributionally robust optimization, last-layer retraining, and related methods show that learned representations and downstream classifiers can play separable roles \citep{sagawa2020distributionally,kirichenko2023last}. Our frozen-representation audits ask how easily a shortcut classifier can be placed on a fixed representation, while our end-to-end experiments test whether early cue reliability changes learned decision rules during training.

\subsection{Architecture, representation, and evaluation}

Vision Transformers \citep{dosovitskiy2021vit}, modern ConvNets \citep{liu2022convnet}, hierarchical transformers \citep{liu2021swin}, self-supervised features \citep{oquab2024dinov2}, and language-supervised vision encoders \citep{radford2021clip,cherti2023reproducible} differ in inductive bias and pretraining scale. We therefore avoid treating CNNs as the only object of study. Our audits include raw pixels, random baselines, ResNet, ConvNeXt, ViT, Swin, DINOv2, and OpenCLIP. The question is not which model is universally best, but whether cue precision predicts shortcut reliance within model families.

Matched accuracy alone is insufficient when the matched test set preserves the shortcut correlation. We therefore prioritize conflict accuracy and texture-choice rate, and use expected calibration error as a secondary diagnostic \citep{guo2017calibration}. Because cue-conflict tests can simplify feature reliance, we add suppression views: texture-suppressed inputs test object recognition when the overlay cue is reduced; object-suppressed inputs test whether texture alone drives texture-class predictions.

\section{Cue-Precision Protocol}

Each example has an object label $y \in \{1,\ldots,K\}$ and a texture label $t \in \{1,\ldots,K\}$. During training or probe fitting, $t=y$ with match probability $p$; otherwise $t$ is sampled from a different class. The evaluation splits are:
\begin{itemize}[leftmargin=2em]
\item \textbf{Matched-ID}: object and texture preserve the training correlation.
\item \textbf{Conflict}: object and texture indicate different classes.
\item \textbf{Texture-suppressed}: object evidence remains while the injected texture cue is reduced or removed.
\item \textbf{Object-suppressed}: texture evidence remains while object evidence is strongly reduced.
\end{itemize}

For a classifier $f$, conflict accuracy is $P[f(x)=y]$. Texture-choice rate is $P[f(x)=t]$ on conflict or object-suppressed examples. We report matched-ID accuracy, conflict accuracy, conflict texture-choice rate, texture-suppressed accuracy, and object-suppressed texture choice as primary or diagnostic endpoints. We retain a descriptive shortcut score only as a secondary summary:
\begin{equation}
\sscore = (1 - \confacc) + \texchoice + \ece_{\mathrm{conflict}}.
\end{equation}
Because the terms partially overlap, conclusions do not rely on this composite metric.

We use two kinds of evidence. In \emph{learning-path} experiments, the training history of the neural model is manipulated directly. In \emph{frozen-probe} audits, the encoder is fixed and cue precision is manipulated during probe fitting; these audits test how easily a shortcut decision boundary can be fitted on a representation, not how the encoder was originally developed.

\section{Experiments}
\label{sec:experiments}

\subsection{Study 1: Synthetic Structural and Decorrelation Curricula}

\subsubsection{Setup}
We first used a synthetic shape-texture task with four shapes and four textures. The target label was shape. Training varied texture-shape match probability. We compared: normal training, degraded-but-predictive input, silhouette-first, mask-first, edge-first, mixed-structural-first, random-texture-first, and a target-schedule control. Models were SmallCNN and MiniViT.

\subsubsection{Results}
Strong shortcut learning emerged in both architectures. In the high-shortcut condition, SmallCNN normal training reached 0.894 matched-ID accuracy but only 0.056 conflict accuracy and 0.879 texture choice. Random-texture-first reached 0.877 matched-ID accuracy, 0.812 conflict accuracy, and 0.102 texture choice. MiniViT normal training reached 1.000 matched-ID accuracy but only 0.038 conflict accuracy and 0.937 texture choice, whereas mixed-structural-first reached 0.643 conflict accuracy and 0.128 texture choice.

\begin{figure}[H]
\centering
\begin{subfigure}{0.48\textwidth}
\includegraphics[width=\linewidth]{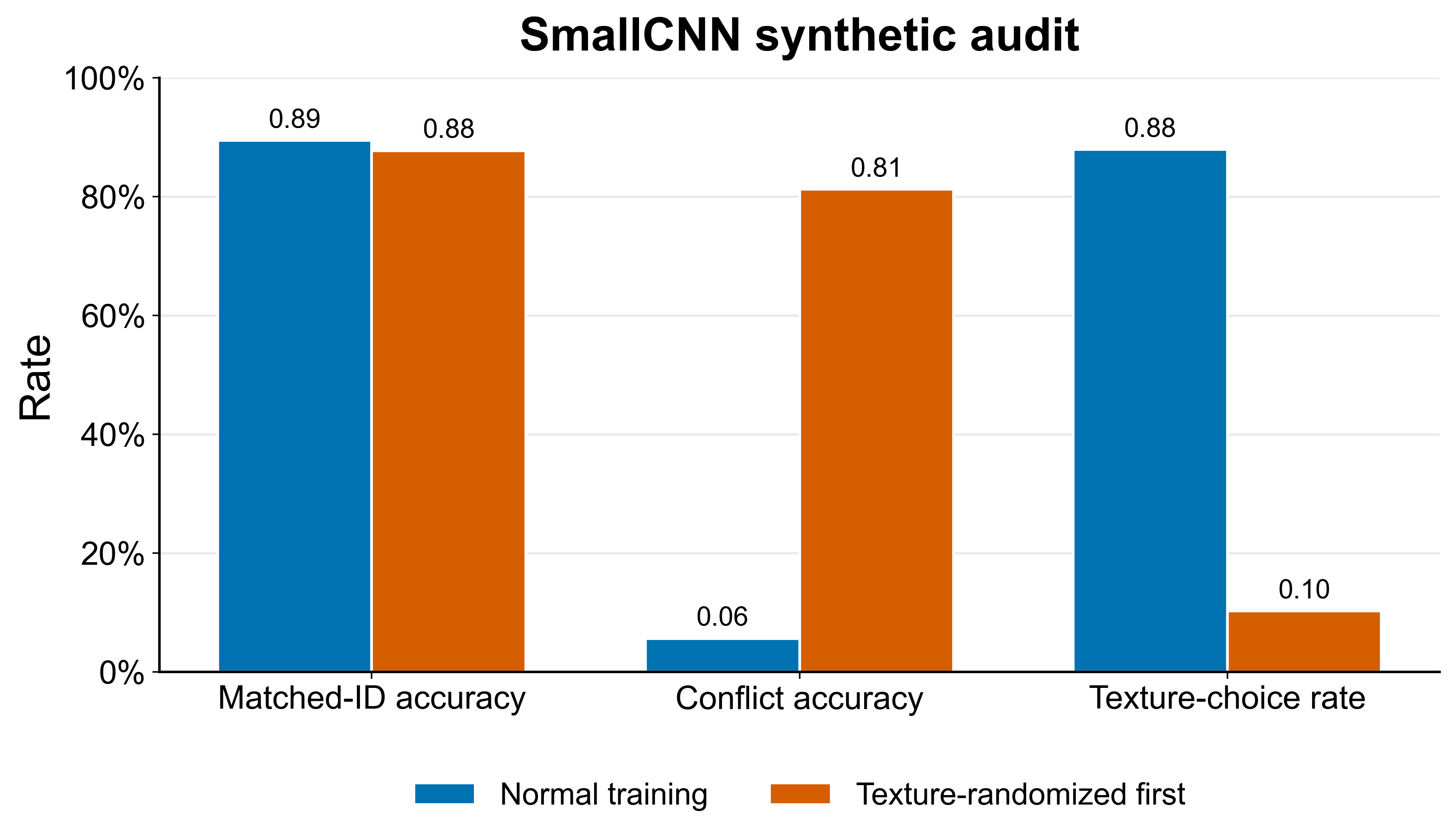}
\caption{SmallCNN}
\end{subfigure}
\hfill
\begin{subfigure}{0.48\textwidth}
\includegraphics[width=\linewidth]{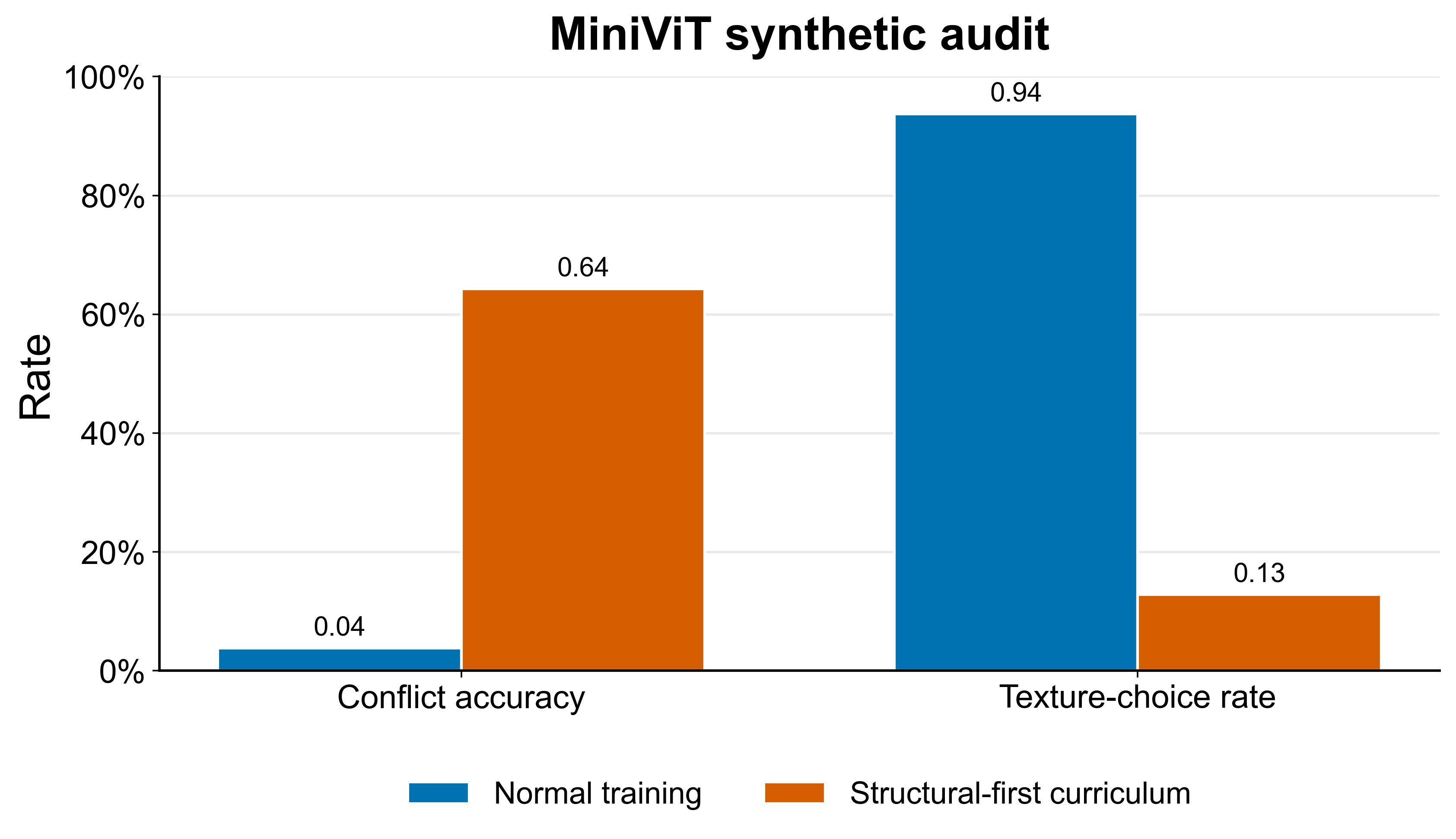}
\caption{MiniViT}
\end{subfigure}
\caption{Synthetic shape-texture curricula. Degraded-but-predictive training is weaker than curricula that reduce early texture predictiveness. Both CNN and ViT-like models learn texture shortcuts when the cue is highly predictive early.}
\label{fig:synthetic}
\end{figure}

These experiments establish two points. Shortcut learning is not specific to CNNs, and visual degradation is not equivalent to reducing the predictive precision of a shortcut cue.

\subsection{Study 2: Sequential Digit Learning}

\subsubsection{Setup}
We next used handwritten digit shapes from scikit-learn \citep{pedregosa2011scikit}. Digit strokes were rendered with class-specific texture patterns. True labels were digit identities. Texture-label match probability was manipulated during pretraining, followed by target-biased fine-tuning. Models were TinyCNN and TinyPatchAttention.

\subsubsection{Results}
The sequential experiments produced a strong dose-response. For TinyCNN after one epoch of target fine-tuning, $p=0.25$ yielded 0.633 conflict accuracy and 0.135 texture choice. At $p=0.75$, conflict accuracy fell to 0.125 and texture choice rose to 0.776. Under target-pretrained $p=1.00$, matched-ID accuracy reached 0.982, but conflict accuracy was only 0.032 and texture choice was 0.901. TinyPatchAttention was more robust overall, but still degraded under perfect early cue precision: $p=0.25$ produced 0.954 conflict accuracy and 0.018 texture choice, whereas $p=1.00$ produced 0.480 conflict accuracy and 0.401 texture choice.

\begin{figure}[H]
\centering
\begin{subfigure}{0.48\textwidth}
\includegraphics[width=\linewidth]{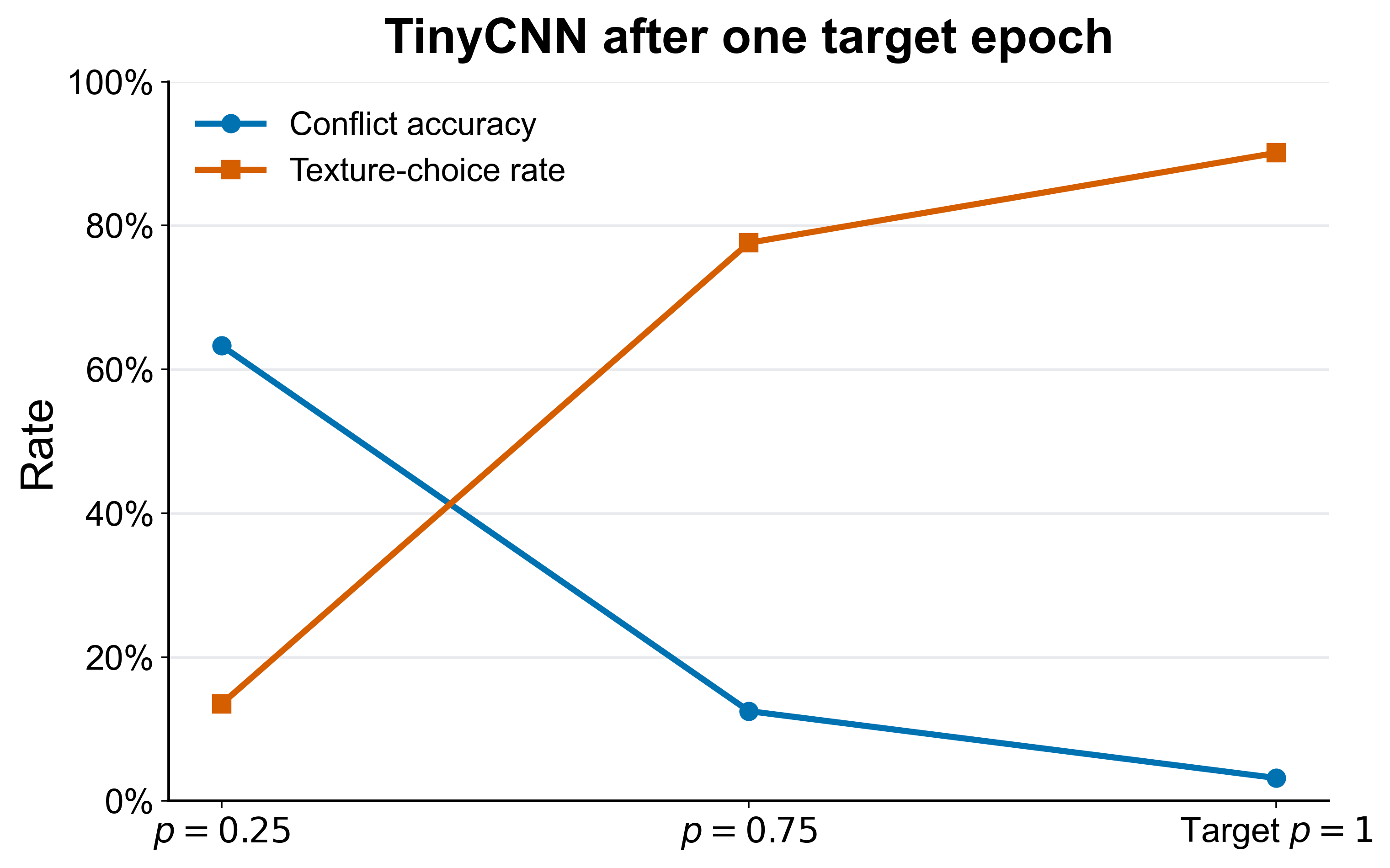}
\caption{TinyCNN}
\end{subfigure}
\hfill
\begin{subfigure}{0.48\textwidth}
\includegraphics[width=\linewidth]{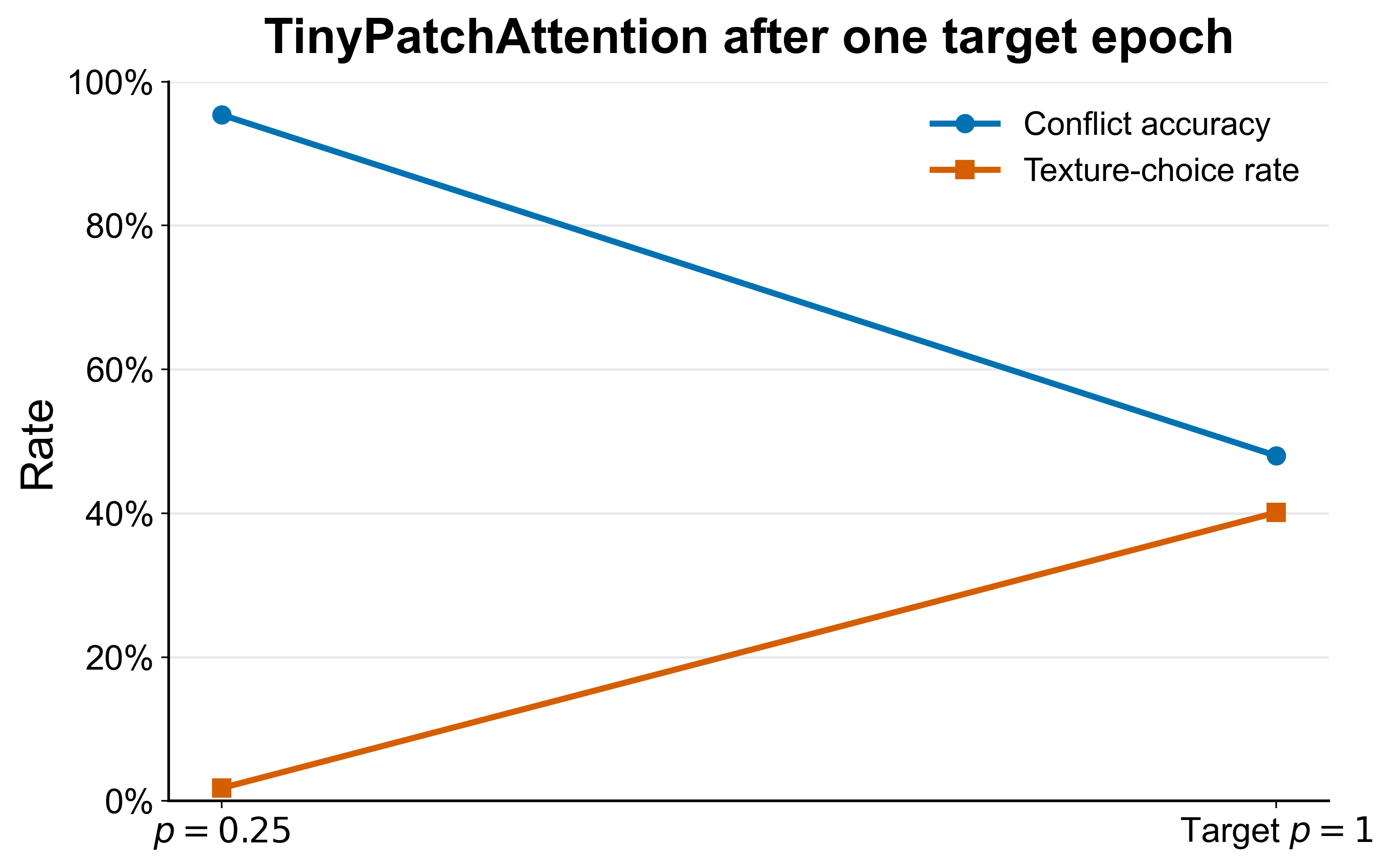}
\caption{TinyPatchAttention}
\end{subfigure}
\caption{True sequential digit learning. Increasing early texture-label precision produces a dose-response decrease in conflict accuracy, although architecture modulates severity.}
\label{fig:digitseq}
\end{figure}

\subsection{Study 3: Frozen Representation Audit on 10-Class Digits}

\subsubsection{Setup}
The third study tested whether cue-biased probes can exploit shortcuts across fixed visual representations. We used a 10-class textured version of the scikit-learn digit dataset. For each seed, train, matched-ID, and conflict base digit indices were disjoint and held fixed across preconditions. We audited raw pixel features, random ResNet50, random ViT-B/16, ResNet18/50, ConvNeXt-Tiny, ViT-B/16, Swin-T, DINOv2 ViT-B/14, and OpenCLIP ViT-B/32. For each frozen representation, a logistic-regression probe predicted digit identity. The primary result uses probe $C=1.0$ and no PCA. Controls used $C \in \{0.01,0.1,1,10\}$, PCA dimensions $\{128,256,512\}$, and color modes including grayscale, fixed palette, class palette, and random palette per image.

\subsubsection{Results}
Table~\ref{tab:digit-audit} shows the primary 10-class result. As early texture-label precision increased, matched-ID accuracy increased, but conflict accuracy decreased and texture choice increased. Target-perfect texture yielded 0.987 matched-ID accuracy, 0.005 conflict accuracy, and 0.975 texture choice. Degraded-but-predictive input remained shortcut-dominated.

\begin{table}[H]
\centering
\caption{10-class digit frozen-representation audit, $C=1$, no PCA, averaged over 10 representations and five seeds. Primary endpoints are conflict accuracy and texture-choice rate.}
\label{tab:digit-audit}
\begin{adjustbox}{max width=\textwidth}
\begin{tabular}{lrrrr}
\toprule
Condition & Matched-ID acc. & Conflict acc. & Texture choice & Score (secondary) \\
\midrule
$p=0.10$ & 0.617 & 0.589 & 0.051 & 0.670 \\
$p=0.25$ & 0.766 & 0.568 & 0.090 & 0.744 \\
$p=0.50$ & 0.861 & 0.505 & 0.174 & 0.939 \\
$p=0.75$ & 0.912 & 0.383 & 0.313 & 1.308 \\
Degraded $p=1$ & 0.873 & 0.038 & 0.796 & 2.558 \\
Target $p=1$ & 0.987 & 0.005 & 0.975 & 2.930 \\
\bottomrule
\end{tabular}
\end{adjustbox}
\end{table}

\begin{figure}[H]
\centering
\includegraphics[width=0.82\textwidth]{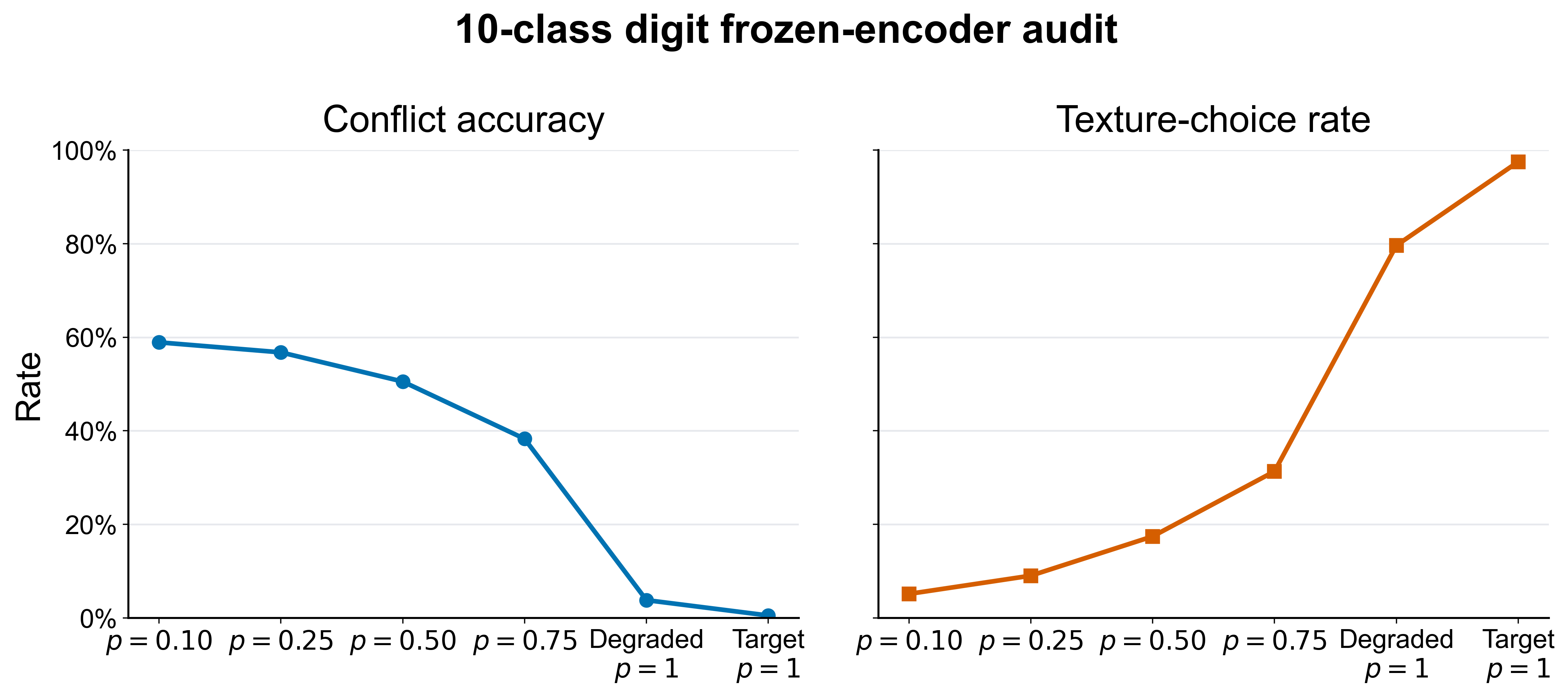}
\caption{10-class digit frozen-representation audit. As early texture-label precision increases, conflict accuracy decreases and texture choice increases.}
\label{fig:digitfrozen}
\end{figure}

Bootstrap slopes confirmed the dose-response. Across the controlled digit audit units, conflict accuracy had slope -0.340 per unit cue precision (95\% CI [-0.353, -0.329]), while texture choice had slope 0.476 (95\% CI [0.460, 0.493]). Paired permutation tests comparing $p=0.10$ to target $p=1.00$ showed a mean conflict-accuracy advantage of 0.571 and texture-choice reduction of 0.926. Permutation tests used 10,000 paired sign/randomization draws, so the minimum non-zero two-sided resolution is approximately $10^{-4}$. Color, PCA, and probe-regularization controls preserved the ordering. Shape decodability strongly predicted low-precision conflict accuracy, supporting the view that cue decorrelation helps when object information remains accessible to the classifier.

\subsection{Study 4: CIFAR-10 Natural-Image-Based Texture-Overlay Benchmark}

\subsubsection{Setup}
To test whether the pattern extends beyond synthetic and digit stimuli, we constructed a CIFAR-10 natural-image-based texture-overlay benchmark \citep{krizhevsky2009learning}. The true label is the CIFAR-10 object class. A class-indexed texture overlay is injected into each image, with $\alpha=0.35$ used as the main overlay strength and $\alpha\in\{0.15,0.25,0.35,0.50\}$ used in a sensitivity sweep. This is a controlled texture-overlay benchmark built on natural-object images, not a claim about naturally occurring CIFAR-10 shortcuts. We evaluated two regimes. In the \emph{frozen-probe} regime, raw pixels, ResNet18/50, ConvNeXt-Tiny, ViT-B/16, DINOv2 ViT-B/14, and OpenCLIP ViT-B/32 features were frozen and a probe was trained under different cue precisions. In the \emph{end-to-end} regime, SmallCNN and ResNet18 were trained sequentially with early cue precision and then exposed to shortcut-rich target fine-tuning for $\ft\in\{0,1,3,6,12,24,48\}$ epochs.

\begin{figure}[H]
\centering
\includegraphics[width=0.96\textwidth]{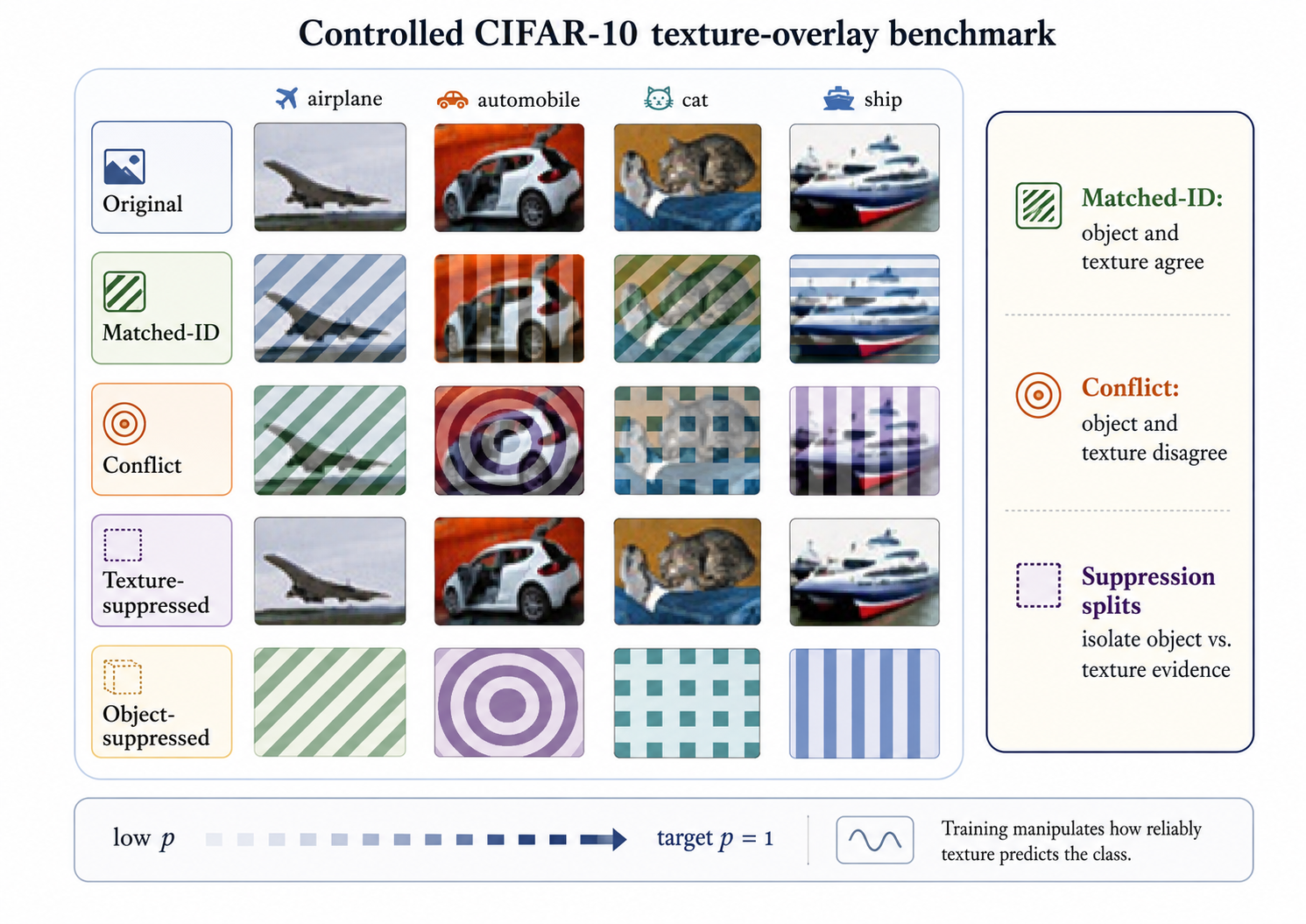}
\caption{Schematic overview of the CIFAR-10 texture-overlay benchmark. Rows show the original image, matched-ID overlay, conflict overlay, texture-suppressed view, and object-suppressed view. The figure is illustrative; the quantitative experiments use the generated benchmark splits described in the text.}
\label{fig:cifar-examples}
\end{figure}

The four CIFAR-10 evaluation views are shown in Fig.~\ref{fig:cifar-examples}. In texture-suppressed examples, the texture overlay is removed and the image is blended only weakly with a neutral field, so object evidence remains. In object-suppressed examples, the object image is heavily degraded by low resolution, blur, and contrast reduction, and then mixed at 10\% with a 90\% texture image. CIFAR-10 has no segmentation masks, so the object-suppressed split is an approximate texture-dominant diagnostic rather than a perfect object/texture disentanglement.

\subsubsection{Frozen-probe results}
CIFAR-10 frozen probes reproduced the dose-response pattern (Table~\ref{tab:cifar-frozen}). Model-averaged conflict accuracy fell from 0.569 at $p=0.10$ to 0.449 at $p=0.75$ and 0.114 at target $p=1.00$. Texture choice rose from 0.049 to 0.293 and then 0.855. The degraded-but-predictive condition was strongly shortcut-dominated: conflict accuracy was 0.056 and texture choice was 0.773. Suppression splits supported the same interpretation: texture-suppressed accuracy was much higher under low-precision training (0.659) than under degraded-but-predictive input (0.129), while object-suppressed texture choice rose with cue precision.

\begin{table}[H]
\centering
\caption{CIFAR-10 frozen-probe audit, model-averaged endpoints across seven representations and three seeds.}
\label{tab:cifar-frozen}
\begin{adjustbox}{max width=\textwidth}
\begin{tabular}{lrrrrr}
\toprule
Condition & Matched-ID acc. & Conflict acc. & Texture choice & Texture-supp. acc. & Object-supp. texture \\
\midrule
$p=0.10$ & 0.579 & 0.569 & 0.049 & 0.659 & 0.107 \\
$p=0.25$ & 0.720 & 0.552 & 0.099 & 0.661 & 0.239 \\
$p=0.50$ & 0.840 & 0.508 & 0.183 & 0.657 & 0.358 \\
$p=0.75$ & 0.913 & 0.449 & 0.293 & 0.627 & 0.480 \\
Degraded $p=1$ & 0.836 & 0.056 & 0.773 & 0.129 & 0.981 \\
Target $p=1$ & 1.000 & 0.114 & 0.855 & 0.356 & 0.837 \\
\bottomrule
\end{tabular}
\end{adjustbox}
\end{table}

\begin{figure}[H]
\centering
\begin{subfigure}{0.48\textwidth}
\includegraphics[width=\linewidth]{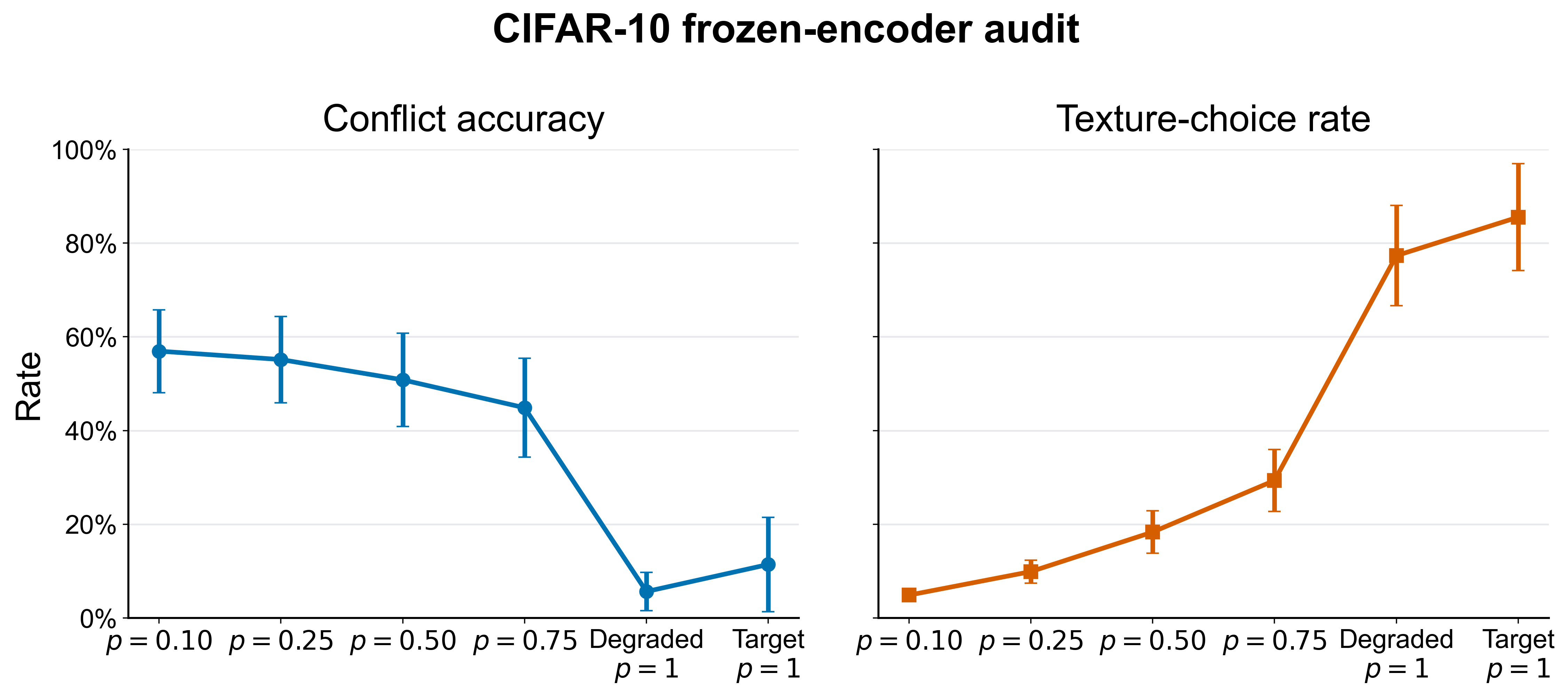}
\caption{Frozen probes}
\end{subfigure}
\hfill
\begin{subfigure}{0.48\textwidth}
\includegraphics[width=\linewidth]{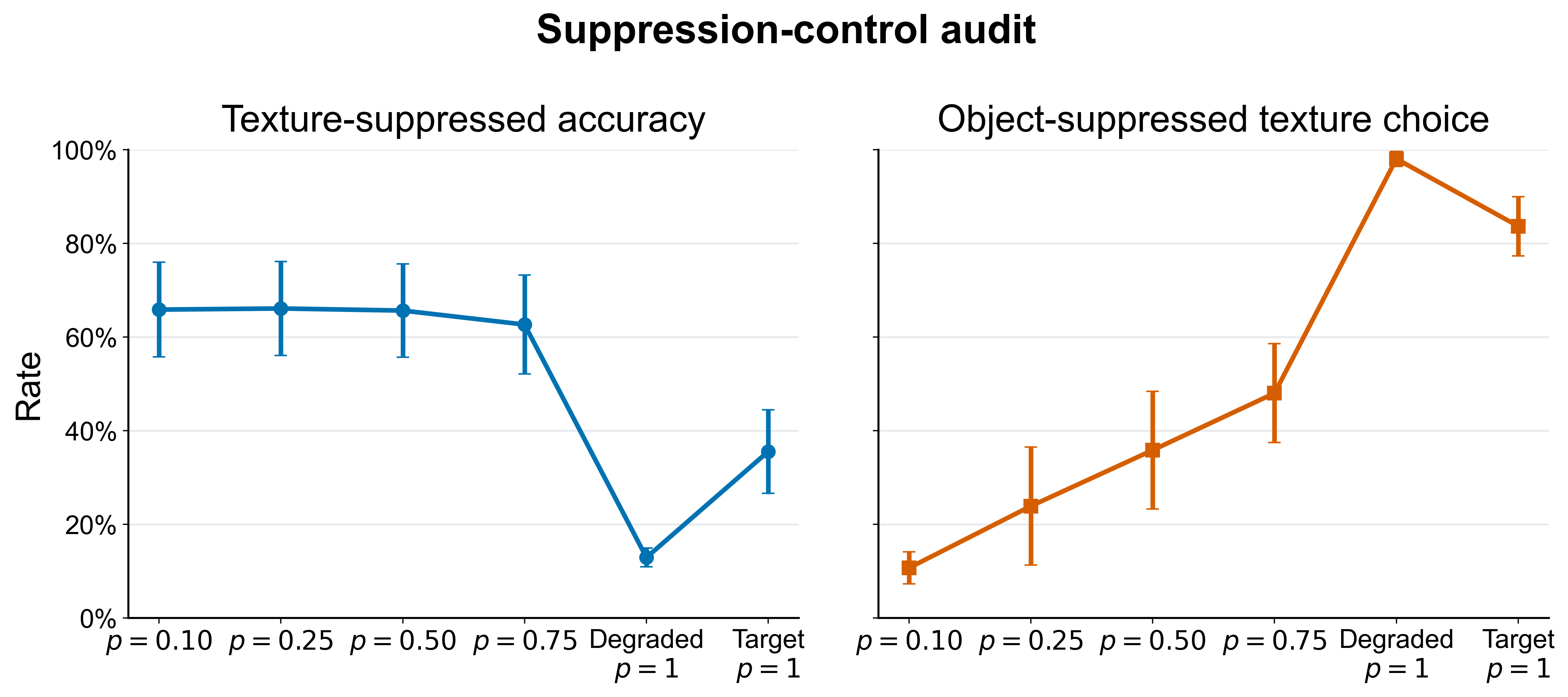}
\caption{Suppression controls}
\end{subfigure}
\caption{CIFAR-10 natural-image-based frozen-probe benchmark. Low cue precision improves conflict and texture-suppressed performance, while high cue precision increases texture choice and object-suppressed texture predictions.}
\label{fig:cifar-frozen}
\end{figure}

The frozen-probe result is not evidence that the pretrained encoders themselves were developmentally trained under our cue-precision conditions. It shows that, on several fixed visual representations, a biased downstream classifier can exploit the most reliable low-level cue. We therefore interpret this component as a representation audit under cue-biased probing.

\subsubsection{End-to-end results and boundary condition}
End-to-end CIFAR-10 training adds a stronger learning-path test. Before target-biased fine-tuning ($\ft=0$), lower early cue precision produced higher conflict accuracy and lower texture choice. Averaged over SmallCNN, ResNet18, and three seeds, $p=0.10$ gave 0.305 conflict accuracy and 0.068 texture choice, whereas $p=0.75$ gave 0.088 conflict accuracy and 0.747 texture choice. Degraded $p=1$ and target $p=1$ were already collapsed.

However, shortcut-rich target fine-tuning rapidly overwrote the low-precision benefit. After one epoch of target-biased fine-tuning, model-averaged conflict accuracy was 0.047 at $p=0.10$ and texture choice was 0.827. By 48 epochs, all end-to-end conditions converged to near-complete texture reliance. The appropriate conclusion is therefore not that low early cue precision permanently prevents shortcut learning, but that it reduces or delays shortcut reliance unless later training reintroduces a highly predictive shortcut.

\begin{table}[H]
\centering
\caption{CIFAR-10 end-to-end learning curve, averaged over SmallCNN, ResNet18, and three seeds. $\ft$ is the number of shortcut-rich target fine-tuning epochs.}
\label{tab:cifar-e2e}
\begin{adjustbox}{max width=\textwidth}
\begin{tabular}{lrrrrrr}
\toprule
Condition & \multicolumn{2}{c}{$\ft=0$} & \multicolumn{2}{c}{$\ft=1$} & \multicolumn{2}{c}{$\ft=48$} \\
\cmidrule(lr){2-3}\cmidrule(lr){4-5}\cmidrule(lr){6-7}
 & Conflict acc. & Texture choice & Conflict acc. & Texture choice & Conflict acc. & Texture choice \\
\midrule
$p=0.10$ & 0.305 & 0.068 & 0.047 & 0.827 & 0.000 & 1.000 \\
$p=0.25$ & 0.243 & 0.290 & 0.017 & 0.941 & 0.000 & 1.000 \\
$p=0.50$ & 0.154 & 0.558 & 0.006 & 0.981 & 0.000 & 1.000 \\
$p=0.75$ & 0.088 & 0.747 & 0.002 & 0.994 & 0.000 & 1.000 \\
Degraded $p=1$ & 0.000 & 0.981 & 0.000 & 0.998 & 0.000 & 1.000 \\
Target $p=1$ & 0.000 & 1.000 & 0.000 & 1.000 & 0.000 & 1.000 \\
\bottomrule
\end{tabular}
\end{adjustbox}
\end{table}

\begin{figure}[H]
\centering
\includegraphics[width=0.92\textwidth]{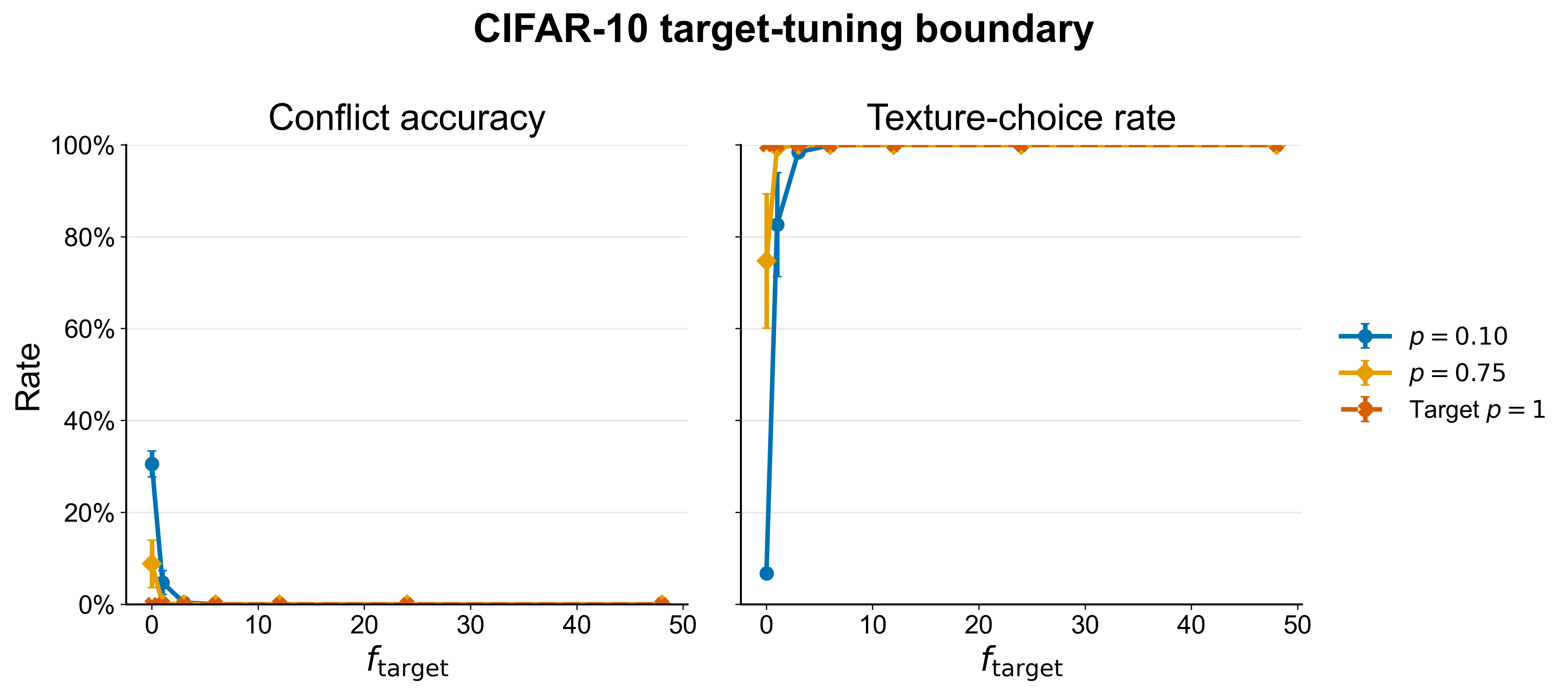}
\caption{CIFAR-10 end-to-end training. Low early cue precision improves conflict behavior before target-biased fine-tuning, but shortcut-rich target exposure rapidly overwrites that benefit.}
\label{fig:cifar-e2e}
\end{figure}

\begin{figure}[H]
\centering
\includegraphics[width=0.98\textwidth]{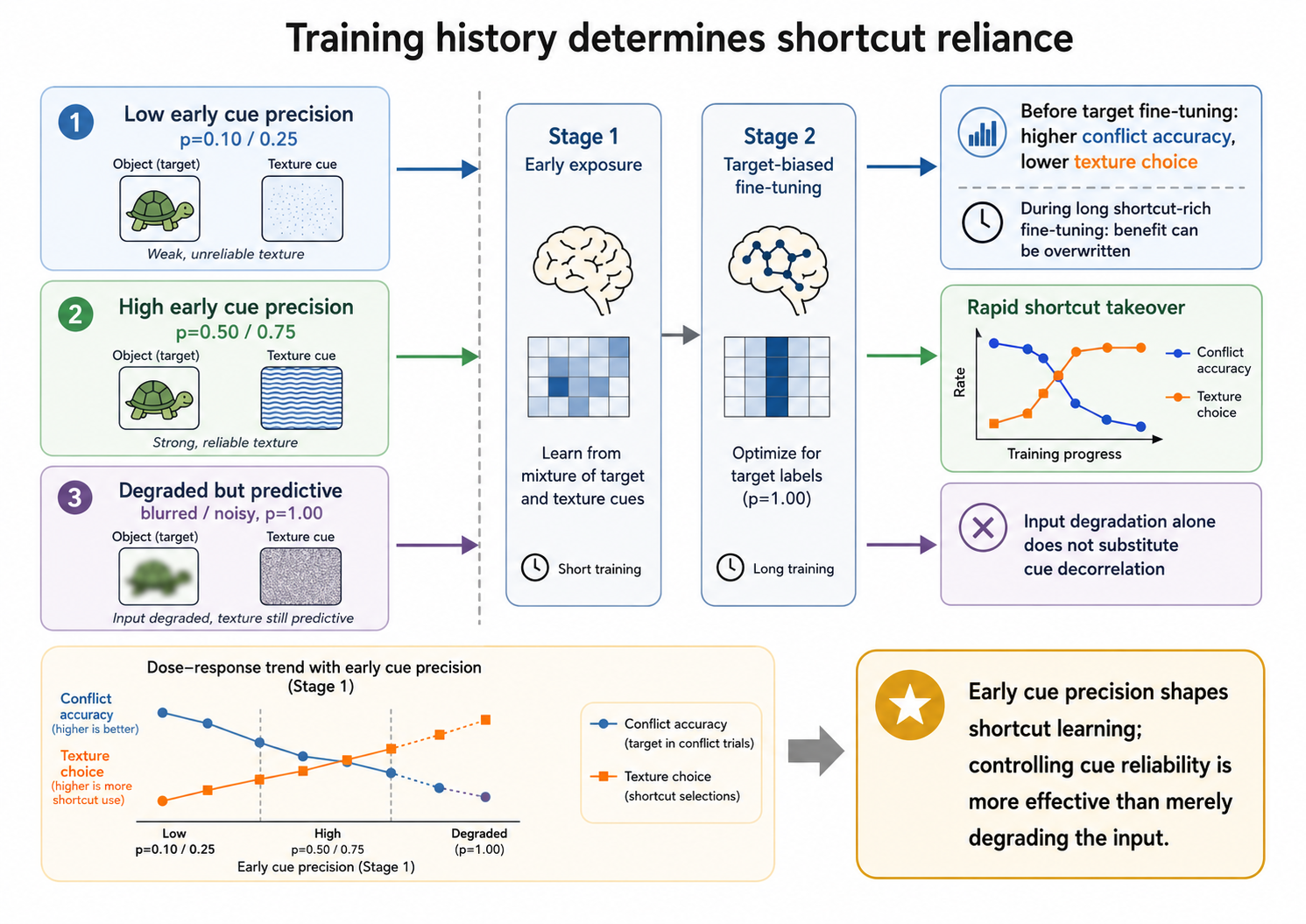}
\caption{Conceptual summary of the end-to-end learning-path result. Low early cue precision improves pre-target conflict behavior, but a later shortcut-rich fine-tuning stage can overwrite that benefit. Degraded-but-predictive input remains distinct from cue decorrelation because the shortcut cue can remain label-predictive despite input degradation.}
\label{fig:training-history-summary}
\end{figure}

Figure~\ref{fig:training-history-summary} summarizes this boundary condition: early cue decorrelation changes the decision rule before target-biased exposure, but robust downstream adaptation requires controlling shortcut reliability throughout fine-tuning.

The CIFAR-10 slopes were statistically consistent with the cue-precision hypothesis. In frozen probes, the conflict-accuracy slope was -0.187 per unit cue precision (95\% CI [-0.216, -0.155]), and texture-choice slope was 0.374 (95\% CI [0.292, 0.461]). A model-clustered bootstrap over frozen-probe model-level slopes produced the same direction: conflict-accuracy slope -0.187 (95\% CI [-0.230, -0.134]) and texture-choice slope 0.374 (95\% CI [0.231, 0.521]). In end-to-end runs across fine-tuning epochs, conflict-accuracy slope was -0.057 (95\% CI [-0.094, -0.025]), and texture-choice slope was 0.184 (95\% CI [0.079, 0.310]). Paired tests comparing $p=0.10$ against target $p=1.00$ used 10,000 permutations and were significant for conflict accuracy and texture choice in both regimes.

\subsubsection{Texture-overlay strength sensitivity}
To address whether the CIFAR-10 pattern depends on the default overlay strength, we repeated the key CIFAR-10 conditions at $\alpha\in\{0.15,0.25,0.35,0.50\}$. The sweep used $p=0.10$, $p=0.75$, and target $p=1.00$ in both frozen-probe and end-to-end regimes. The frozen-probe ordering was stable across all overlay strengths (Table~\ref{tab:alpha-sensitivity}). Conflict accuracy was always highest at $p=0.10$, lower at $p=0.75$, and lowest under target $p=1.00$; texture-choice rate showed the opposite ordering. As expected, larger $\alpha$ made texture evidence more dominant, but it did not change the qualitative conclusion. For example, at $\alpha=0.15$, conflict accuracy was 0.704 under $p=0.10$ and 0.282 under target $p=1.00$; at $\alpha=0.50$, the corresponding values were 0.520 and 0.078. End-to-end sensitivity runs showed the same pre-target ordering, but also reinforced the boundary condition: shortcut-rich fine-tuning rapidly reduced the low-precision advantage. Full alpha-sensitivity curves and slope intervals are reported in the supplement.

\begin{table}[H]
\centering
\caption{CIFAR-10 frozen-probe overlay-strength sensitivity. Values are model-averaged endpoints over seven representations and three seeds for the subset used in the alpha sweep.}
\label{tab:alpha-sensitivity}
\begin{adjustbox}{max width=\textwidth}
\begin{tabular}{rrrrrrr}
\toprule
$\alpha$ & \multicolumn{3}{c}{Conflict accuracy} & \multicolumn{3}{c}{Texture choice} \\
\cmidrule(lr){2-4}\cmidrule(lr){5-7}
 & $p=0.10$ & $p=0.75$ & Target $p=1$ & $p=0.10$ & $p=0.75$ & Target $p=1$ \\
\midrule
0.15 & 0.704 & 0.621 & 0.282 & 0.033 & 0.189 & 0.603 \\
0.25 & 0.622 & 0.517 & 0.158 & 0.043 & 0.249 & 0.786 \\
0.35 & 0.569 & 0.449 & 0.114 & 0.049 & 0.293 & 0.855 \\
0.50 & 0.520 & 0.383 & 0.078 & 0.056 & 0.343 & 0.903 \\
\bottomrule
\end{tabular}
\end{adjustbox}
\end{table}

\section{Controls and Robustness Checks}

\subsection{Color, PCA, and probe regularization}
Color controls on the 4-class audit showed that the result is not reducible to class-specific color. With grayscale textures, conflict accuracy was 0.814 under $p=0.25$ but 0.026 under target $p=1.00$; with random palettes per image, the corresponding values were 0.826 and 0.046. PCA controls on the 10-class digit audit also preserved the trend. At PCA dimension 128, conflict accuracy was 0.551 for $p=0.10$, 0.375 for $p=0.75$, and 0.005 for target $p=1.00$. Probe-regularization sweeps over $C=0.01,0.1,1,10$ maintained the same ordering.

\subsection{Decodability, per-model trajectories, and model heterogeneity}
The strongest representation-level predictor was shape decodability. In the 10-class digit audit, texture decodability was high across representations, but shape decodability varied substantially. Low-precision conflict accuracy tracked whether shape information was linearly accessible. This suggests that cue decorrelation is most effective when the representation preserves object information that the downstream classifier can use.

We report per-model CIFAR-10 frozen-probe trajectories in the supplement (Figs.~S2--S3). Most representations show the same within-model trend: conflict accuracy declines and texture choice rises as cue precision increases. Model rankings should not be overinterpreted: raw pixels and random features can perform strongly on simpler benchmarks, and CIFAR-10 features differ in pretraining objective and dimensionality. The robust conclusion is the within-model cue-precision trend, not a universal architecture ranking.

\subsection{Accuracy-matched comparisons}
A possible concern is that low-precision conditions are simply underfit. Accuracy-matched comparisons reduce this concern in the frozen-probe setting: pairs with similar matched-ID accuracy can still show better conflict behavior under lower cue precision. For example, in CIFAR-10 frozen probes, ViT-B/16 at $p=0.50$ and degraded $p=1.00$ had nearly identical matched-ID accuracy (0.894 vs. 0.891) but very different conflict accuracy (0.725 vs. 0.037) and texture choice (0.089 vs. 0.851). ResNet50 at $p=0.25$ and degraded $p=1.00$ similarly matched ID accuracy within 0.012 or less across three seeds but differed by roughly 0.38--0.40 in conflict accuracy. These examples show that conflict robustness is not explained solely by lower matched-ID accuracy. In end-to-end CIFAR-10, the opposite limitation appears after target fine-tuning: once matched-ID accuracy saturates near 1.0 for all conditions, conflict behavior also collapses, indicating that strongly biased downstream exposure can erase early benefits.

\section{Discussion}

Across controlled synthetic tasks, sequential digit learning, frozen representation audits, and CIFAR-10 natural-image-based texture overlays, the same qualitative pattern appears. When texture is highly predictive early, models achieve high matched-ID accuracy but often fail under conflict and suppression. When the same cue is visible but less predictive, conflict behavior improves. Degraded-but-predictive input is not enough: if the low-level cue still predicts the label, models often learn the shortcut.

These results support a learning-history view of shortcut learning. The central issue is not whether a cue is present, nor whether an architecture is convolutional, but whether a cue gains excessive reliability during the phase in which a decision rule forms. This view is compatible with prior work emphasizing data and augmentation as sources of texture bias \citep{hermann2020origins}. It also clarifies why simple degradation can fail: degradation changes visual quality but may leave the shortcut's label predictiveness intact.

The end-to-end CIFAR-10 result is especially important because it limits the claim. Low early cue precision reduces or delays shortcut reliance before target-biased exposure, but shortcut-rich fine-tuning can rapidly overwrite the benefit. Thus, cue decorrelation should not be treated as a one-time inoculation. Robust downstream adaptation may require keeping shortcut predictiveness controlled throughout fine-tuning, or combining cue decorrelation with objectives that preserve object-structural information.

\section{Conclusion}

Our controlled benchmarks support an early cue-precision account of visual shortcut learning. Low-level texture cues that are highly predictive early lead to high matched-ID accuracy but poor conflict and suppression behavior. Cue decorrelation reduces or delays shortcut reliance more effectively than degraded-but-predictive input, but the benefit is bounded by representation quality and can be overwritten by later shortcut-rich fine-tuning. These conclusions are intentionally limited to controlled cue manipulations: the CIFAR-10 study uses artificial texture overlays, and frozen-probe audits measure downstream classifier behavior on fixed representations rather than the original pretraining histories of the encoders. Robust visual learning should therefore control the reliability of shortcut cues during both early learning and downstream adaptation, not only alter architecture or degrade visual input quality.

\section*{Data and Code Availability}
Supplementary materials are currently available at https://github.com/xvega123/Early-Cue-Precision-Shapes-Visual-Shortcut-Learning-in-Controlled-Cue-Manipulation-Benchmarks. The complete raw-result and code archive will be deposited in a permanent public repository upon acceptance of the manuscript.

\section*{Ethics Statement}
This study uses synthetic stimuli, scikit-learn handwritten digit data, and CIFAR-10 images. It does not involve human subjects, animal subjects, patient data, or personal data.

\section*{Funding}
This research did not receive any specific grant from funding agencies in the public, commercial, or not-for-profit sectors.

\section*{Declaration of Competing Interest}
The authors declare no competing interests.

\section*{Declaration of Generative AI and AI-Assisted Technologies in the Writing Process}
During the preparation of this work, the authors used AI tools to assist with code drafting, language editing, manuscript organization, and figure generation. The authors thoroughly reviewed all AI-assisted content and take full responsibility for the final publication.

\bibliographystyle{elsarticle-harv}
\bibliography{references}

\end{document}